%% file: main.tex
\pdfoutput=1
\documentclass[runningheads]{llncs}

\usepackage[utf8]{inputenc}
\usepackage[T1]{fontenc}

\usepackage{graphicx}
\usepackage{amsmath,amssymb}
\usepackage{hyperref}

\usepackage{booktabs}
\usepackage{acronym}
\usepackage{csquotes}
\usepackage{tikz}
\usetikzlibrary{arrows,arrows.meta,automata,backgrounds,calc,patterns,positioning,shapes,shadows}
\usepackage[linesnumbered, ruled, noend]{algorithm2e}
\usepackage[capitalise]{cleveref}
\usepackage{acronym}
\usepackage{subcaption}
\input{defs.tex}

\title{Towards Privacy-Preserving Relational Data Synthesis via Probabilistic Relational Models}
\titlerunning{Towards Privacy-Preserving Relational Data Synthesis via PRMs}

\author{
	Malte Luttermann\inst{1} \and
	Ralf Möller\inst{2} \and
	Mattis Hartwig\inst{1,3}
}
\institute{
	German Research Center for Artificial Intelligence (DFKI), Lübeck \\
	\email{\{malte.luttermann,mattis.hartwig\}@dfki.de}
	\and
	Institute for Humanities-Centered Artificial Intelligence, University of Hamburg \\
	\email{ralf.moeller@uni-hamburg.de}
	\and
	singularIT GmbH, Leipzig \\
}

\begin{document}

\maketitle

\begin{abstract}
	Probabilistic relational models provide a well-established formalism to combine first-order logic and probabilistic models, thereby allowing to represent relationships between objects in a relational domain.
	At the same time, the field of artificial intelligence requires increasingly large amounts of relational training data for various machine learning tasks.
	Collecting real-world data, however, is often challenging due to privacy concerns, data protection regulations, high costs, and so on.
	To mitigate these challenges, the generation of synthetic data is a promising approach.
	In this paper, we solve the problem of generating synthetic relational data via probabilistic relational models.
	In particular, we propose a fully-fledged pipeline to go from relational database to probabilistic relational model, which can then be used to sample new synthetic relational data points from its underlying probability distribution.
	As part of our proposed pipeline, we introduce a learning algorithm to construct a probabilistic relational model from a given relational database.
\end{abstract}
\begin{keywords}
	probabilistic graphical models; relational models; synthetic data.
\end{keywords}


\section{Introduction}
Probabilistic relational models such as \acp{pfg} combine first-order logic and probabilistic models and thereby provide an adequate formalism to represent relationships between objects in a relational domain.
\Acp{pfg} compactly encode a full joint probability distribution over a set of \acp{rv} and hence allow to sample new relational data points from the encoded underlying full joint probability distribution.
The generated synthetic relational data samples then follow the underlying full joint probability distribution and might be used for various purposes.
Common applications of synthetic data include for example training machine learning models or sharing data without violating the privacy of individuals~\cite{Chen2021a,Nikolenko2021a,Yoon2019a}.
Another application could be to bootstrap \ac{pfg} learning (using the given database to learn a \ac{pfg}, which is then applied to generate additional synthetic relational data points to learn another \ac{pfg} on the basis of a larger data set and then possibly repeating the procedure).
In this paper, we solve the problem of applying probabilistic relational models (more specifically, \acp{pfg}) to generate synthetic relational (i.e., multi-table) data from a conceptual point of view.

\paragraph{Previous Work.}
While there is, to the best of our knowledge, no previous work on the generation of synthetic relational data via probabilistic relational models, there exists previous work on learning and sampling models that combine probabilistic models and first-order logic.
In particular, so-called \acp{mln}~\cite{Richardson2006a} have extensively been studied.
An \ac{mln} is another formalism combining probabilistic models and first-order logic, allowing for probabilistic reasoning on a first-order level.
Even though a vast amount of algorithms to learn an \ac{mln} from data has emerged (see, e.g., \cite{Biba2008b,Biba2008a,Khot2011a,Kok2005a,Kok2009a,Kok2010a,Lowd2007a,Mihalkova2007a,Singla2005a}), there is a lack of prior work on generating synthetic relational data via \acp{mln} or other probabilistic relational models.
Moreover, while sampling algorithms for \acp{mln} have been developed~\cite{Venugopal2014a}, these sampling algorithms are used for (approximate) query answering and are not applied to synthetic data generation, where the requirements of the sampling algorithm slightly differ---that is, drawing a new data sample should yield a value for every column in the given database.

Most of the existing work on synthetic data generation focuses on the generation of single-table synthetic data~\cite{Figueira2022a}.
While many popular approaches are based on generative adversarial networks~\cite{Xu2019a}, there are also approaches building on probabilistic graphical models such as \aclp{bn}~\cite{Gogoshin2021a}.
Both for approaches based on generative adversarial networks and probabilistic graphical models, differential privacy guarantees have been investigated~\cite{Bao2021a,Cai2021a,Fang2022a,Yoon2019a}.
More recently, there has also been work on the generation of synthetic relational data under differential privacy~\cite{Cai2023a}.
However, none of these approaches makes use of the advantages of first-order probabilistic models.

\paragraph{Our Contributions.}
In this paper, we introduce the first architecture that combines \acp{pfg} and the generation of synthetic relational data.
Our main contribution is a fully-fledged pipeline from relational database to \ac{pfg}, from which we can then sample new realistic synthetic data points.
We further present a learning algorithm to obtain a \ac{pfg} (that is, both the graph structure and the parameters of the \ac{pfg}) from a relational database.
Our proposed architecture exploits the advantages of \acp{pfg}, including that (i) \acp{pfg} are able to effectively encode the relationships between objects in a relational domain, (ii) \acp{pfg} provide an explainable model that can also be used for, e.g., probabilistic inference~\cite{Taghipour2013a} and causal inference~\cite{Luttermann2024b} on a first-order level, and (iii) \acp{pfg} naturally abstract from individuals by grouping indistinguishable objects, which yields a promising foundation for differential privacy guarantees.

\paragraph{Structure of this Paper.}
The remainder of this paper is structured as follows.
We begin by introducing necessary background information and notations.
Afterwards, we propose an architecture to first learn a \ac{pfg} from a relational database and then employ the \ac{pfg} to generate new synthetic relational data points, which might be used for arbitrary applications.
We then go through the individual steps of the proposed architecture in detail, using a small comprehensive example for guidance, before we conclude.

\section{Preliminaries} \label{sec:syn_prelims}
We start with the definition of \acp{fg} as propositional probabilistic models and then continue to introduce \acp{pfg}, which combine first-order logic with probabilistic models.
An \ac{fg} is an undirected propositional probabilistic model to compactly encode a full joint probability distribution over a set of \acp{rv}~\cite{Frey1997a,Kschischang2001a}.
Similar to a \acl{bn}~\cite{Pearl1988a}, an \ac{fg} factorises a full joint probability distribution into a product of factors.
\begin{definition}[Factor Graph]
	An \emph{\ac{fg}} $G = (\boldsymbol V, \boldsymbol E)$ is an undirected bipartite graph with node set $\boldsymbol V = \boldsymbol R \cup \boldsymbol \Phi$ where $\boldsymbol R = \{R_1, \ldots, R_n\}$ is a set of variable nodes (\acp{rv}) and $\boldsymbol \Phi = \{\phi_1, \ldots, \phi_m\}$ is a set of factor nodes (functions).
	The term $\range{R_i}$ denotes the possible values of a \ac{rv} $R_i$.
	There is an edge between a variable node $R_i$ and a factor node $\phi_j$ in $\boldsymbol E \subseteq \boldsymbol R \times \boldsymbol \Phi$ if $R_i$ appears in the argument list of $\phi_j$.
	A factor is a function that maps its arguments to a positive real number, called potential.
	The semantics of $G$ is given by
	$$
		P_G = \frac{1}{Z} \prod_{j=1}^m \phi_j(\mathcal A_j)
	$$
	with $Z$ being the normalisation constant and $\mathcal A_j$ denoting the \acp{rv} connected to $\phi_j$ (that is, the arguments of $\phi_j$).
\end{definition}
\begin{figure}[t]
	\centering
	\resizebox{\textwidth}{!}{\input{files/example_fg.tex}}
	\caption{An \ac{fg} encoding a full joint probability distribution for an epidemic example~\cite{Hoffmann2022a}. We omit the input-output pairs of the factors for brevity.}
	\label{fig:example_fg}
\end{figure}
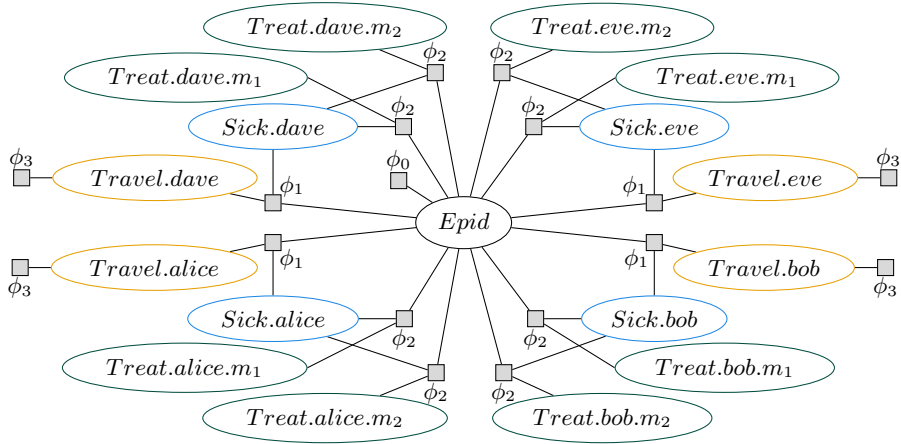
\begin{example} \label{ex:fg_example}
	Take a look at the \ac{fg} presented in \cref{fig:example_fg}, which represents an epidemic example with four persons $alice$, $bob$, $dave$, and $eve$ as well as two possible medications $m_1$ and $m_2$ for treatment.
	For each person, there are two Boolean \acp{rv} (that is, \acp{rv} having a Boolean range) $Sick$ and $Travel$, indicating whether the person is sick and travels, respectively.
	Moreover, there is another Boolean \ac{rv} $Treat$ for each combination of person and medication, specifying whether the person is treated with the medication.
	The Boolean \ac{rv} $Epid$ states whether an epidemic is present.
\end{example}
Next, we define \acp{pfg}, first introduced by Poole~\cite{Poole2003a}, which combine probabilistic models and first-order logic.
\Acp{pfg} use \acp{lv} as parameters to represent sets of indistinguishable \acp{rv}.
Each set of indistinguishable \acp{rv} is represented by a \ac{prv}.
\begin{definition}[Parameterised Random Variable]
	Let $\boldsymbol{R}$ be a set of \ac{rv} names, $\boldsymbol{L}$ a set of \ac{lv} names, and $\boldsymbol{D}$ a set of constants.
	All sets are finite.
	Each \ac{lv} $L$ has a domain $\domain{L} \subseteq \boldsymbol{D}$.
	A \emph{constraint} is a tuple $(\mathcal{X}, C_{\mathcal{X}})$ of a sequence of \acp{lv} $\mathcal{X} = (X_1, \ldots, X_n)$ and a set $C_{\mathcal{X}} \subseteq \times_{i = 1}^n \domain{X_i}$.
	The symbol $\top$ for $C$ marks that no restrictions apply, i.e., $C_{\mathcal{X}} = \times_{i = 1}^n \domain{X_i}$.
	A \emph{\ac{prv}} $R(L_1, \ldots, L_n)$, $n \geq 0$, is a syntactical construct of a \ac{rv} $R \in \boldsymbol{R}$ possibly combined with \acp{lv} $L_1, \ldots, L_n \in \boldsymbol{L}$ to represent a set of \acp{rv}.
	If $n = 0$, the \ac{prv} is parameterless and forms a propositional \ac{rv}.
	A \ac{prv} $A$ (or \ac{lv} $L$) under constraint $C$ is given by $A_{|C}$ ($L_{|C}$), respectively.
	We may omit $|\top$ in $A_{|\top}$ or $L_{|\top}$.
	The term $\range{A}$ denotes the possible values of a \ac{prv} $A$. 
	An \emph{event} $A = a$ denotes the occurrence of \ac{prv} $A$ with range value $a \in \range{A}$.
\end{definition}
\begin{example} \label{ex:prv_example}
	Consider $\boldsymbol{R} = \{Epid, Travel, Sick, Treat\}$ and $\boldsymbol{L} = \{P,M\}$ with $\domain{P} = \{alice, bob, dave, eve\}$ (patients), $\domain{M} = \{m_1, m_2\}$ (medications), combined into Boolean \acp{prv} $Epid$, $Travel(P)$, $Sick(P)$, and $Treat(P,M)$.
\end{example}
A \ac{pf} describes a function, mapping argument values to positive real numbers, of which at least one is non-zero.
\begin{definition}[Parfactor]
	Let $\Phi$ denote a set of factor names.
	We denote a \emph{\ac{pf}} $g$ by $\phi(\mathcal{A})_{| C}$ with $\mathcal{A} = (A_1, \ldots, A_n)$ being a sequence of \acp{prv}, $\phi$$:$ $\times_{i = 1}^n \range{A_i} \mapsto \mathbb{R}^+$ being a function with name $\phi \in \Phi$ mapping argument values to a positive real number called \emph{potential}, and $C$ being a constraint on the \acp{lv} of $\mathcal{A}$.
	We may omit $|\top$ in $\phi(\mathcal{A})_{|\top}$.
	The term $lv(Y)$ refers to the \acp{lv} in some element $Y$, a \ac{prv}, a \ac{pf}, or sets thereof.
	The term $gr(Y_{| C})$ denotes the set of all instances (groundings) of $Y$ with respect to constraint $C$.
\end{definition}
A \ac{pfg} is then built from a set of \acp{pf} $\{g_1, \dots, g_m\}$.
\begin{definition}[Parametric Factor Graph]
	A \emph{\ac{pfg}} $G = (\boldsymbol V, \boldsymbol E)$ is a bipartite graph with node set $\boldsymbol V = \boldsymbol A \cup \boldsymbol G$ where $\boldsymbol A = \{A_1, \ldots, A_n\}$ is a set of \acp{prv} and $\boldsymbol G = \{g_1, \ldots, g_m\}$ is a set of \acp{pf} as well as edge set $\boldsymbol E \subseteq \boldsymbol A \times \boldsymbol G$.
	A \ac{prv} $A_i$ and a \ac{pf} $g_j$ are connected via an edge in $G$ (i.e., $\{A_i, g_j\} \in \boldsymbol E$) if $A_i$ appears in the argument list of $g_j$.
	The semantics of $G$ is given by grounding and building a full joint distribution.
	With $Z$ as the normalisation constant and $\mathcal A_k$ denoting the \acp{rv} connected to $\phi_k$, $G$ represents the full joint distribution
	$$
		P_G = \frac{1}{Z} \prod_{g_j \in \boldsymbol G} \prod_{\phi_k \in gr(g_j)} \phi_k(\mathcal A_k).
	$$
\end{definition}
\begin{example}
	\Cref{fig:example_pfg} shows a \ac{pfg} $G = \{g_i\}^3_{i=0}$ with $g_0 = \phi_0(Epid)_{| \top}$, $g_1 = \phi_1(Travel(P),\allowbreak Sick(P),\allowbreak Epid)_{| \top}$, $g_2 = \phi_2(Treat(P,M),\allowbreak Sick(P),\allowbreak Epid)_{| \top}$, and $g_3 = \phi_2(Travel(P))_{| \top}$.
	Grounding $G$ yields again the \ac{fg} shown in \cref{fig:example_fg} (assuming that the domains of the \acp{lv} are defined as in \cref{ex:prv_example}).
\end{example}
\begin{figure}[t]
	\centering
	\input{files/example_pfg.tex}
	\caption{A \ac{pfg} encoding a full joint probability distribution for the epidemic example from \cref{fig:example_fg}. We omit the input-output pairs of the \acp{pf} for brevity.}
	\label{fig:example_pfg}
\end{figure}
Note that the definition of \acp{pfg} also includes \acp{fg}, as every \ac{fg} is a \ac{pfg} containing only parameterless \acp{rv}.
Compared to an \ac{fg}, a \ac{pfg} abstracts from individuals by grouping identically behaving objects using \acp{lv}.
While the introduction of \acp{lv} increases the expressiveness of the model (e.g., to encode relationships between groups of objects), grouping identically behaving individuals can also yield significant speed-ups during probabilistic inference.

In the upcoming section, we introduce our proposed architecture to solve the problem of learning a \ac{pfg} from a given relational database to then generate synthetic relational data according to its underlying probability distribution.

\section{Proposed Architecture}
In this section, we provide an overview of the general architecture to generate synthetic relational data from a relational database using a \ac{pfg}.
We first take a look at the steps involved in the synthetic data generation approach and afterwards continue to investigate each of the steps in more detail.

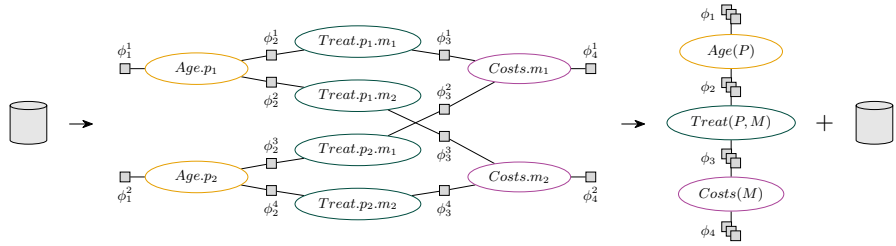
\begin{figure}[t]
	\centering
	\input{files/architecture_overview.tex}
	\caption{Architecture overview of the general architecture to generate synthetic relational data from a given relational database using a \ac{pfg}.}
	\label{fig:architecture_overview}
\end{figure}

An overview of the architecture of our proposed architecture is depicted in \cref{fig:architecture_overview}.
The whole process consists of three primary steps, which can again be decomposed into various subroutines.
The three primary steps consist of (i) constructing a propositional \ac{fg}, (ii) transforming the propositional \ac{fg} into a \ac{pfg}, and (iii) sampling from the \ac{pfg} to generate new synthetic relational data.
Besides the generated synthetic data, the \ac{pfg} is also a valuable output of the architecture (indicated by the $+$ sign), as it can be used for various tasks such as probabilistic inference, causal inference, or bootstrap \ac{pfg} learning, for example.

In the following, we illustrate each of the three steps at a small comprehensive example for guidance.
Before we take a look at the steps involved in the proposed architecture, we briefly introduce the notion of an \ac{er} model, which allows us to describe a relational database $\mathcal D$.
\begin{definition}[Entity-Relationship Model]
	We define an \emph{\ac{er} model} as a tuple $(\mathcal E, \mathcal R)$ where $\mathcal E = \{E_1, \ldots, E_{\ell}\}$ denotes a set of entity classes and $\mathcal R = \{R_1, \ldots, R_k\}$ is a set of relationship classes.
	Each entity or relationship class $B \in \mathcal E \cup \mathcal R$ can have a set of attributes attached to it, which is denoted by $\mathcal A(B)$.
\end{definition}
\begin{example}
	Consider the \ac{er} model depicted in \cref{fig:er_example}, which consists of the entity classes $\mathcal C = \{Patient, Medication\}$ and the relationship classes $\mathcal R = \{Treat\}$.
	The \ac{er} model further contains the attributes $\mathcal A(Patient) = \{Age\}$ and $\mathcal A(Medication) = \{Costs\}$.
\end{example}
It is generally possible to have cyclic relationships and to impose cardinality constraints on the relationships, which we omit for brevity in this paper.
We next take a closer look at the individual steps involved in our proposed architecture.

\subsection{Construction of a Propositional Factor Graph}
While there are learning algorithms for first-order probabilistic models such as \acp{mln}, to the best of our knowledge, there is currently no approach to directly learn a \ac{pfg} from a given relational database. 
However, there are well-known approaches to learn an \ac{fg} from the given data~\cite{Abbeel2006a} and an \ac{fg} can be transformed into a \ac{pfg} by running the so-called \ac{acp} algorithm~\cite{Luttermann2024a}.
We therefore propose to first learn a propositional model, that is, an \ac{fg} $G$, from the given relational database and afterwards run the \ac{acp} algorithm on $G$ to transform $G$ into a \ac{pfg} entailing equivalent semantics as $G$.

While such an approach seems straightforward at first glance, there are a few challenges to overcome.
A major challenge is that applying a standard learning algorithm to obtain an \ac{fg} from data does not fit our setting because standard learning algorithms do not include \acp{rv} and factors for individual objects into the \ac{fg}.
In other words, the relational structure of the data is neglected, which we illustrate in the upcoming example.
\begin{figure}[t]
	\centering
	\begin{subfigure}[t]{\linewidth}
		\centering
		\input{files/er_example.tex}
		\caption{}
		\label{fig:er_example}
	\end{subfigure}

	\begin{subfigure}[t]{\linewidth}
		\centering
		\input{files/db_example.tex}
		\caption{}
		\label{fig:db_table_example}
	\end{subfigure}
	\caption{A small toy example for a relational database of patients and the medications they take. \Cref{fig:er_example} shows an \ac{er} model consisting of entities $Patient$ and $Medication$ with attributes $Age$ and $Costs$, respectively.
	The entities $Patient$ and $Medication$ are connected by a relation $Treat$. We omit cardinalities for brevity. \Cref{fig:db_table_example} displays exemplary data for a relational database following the structure specified by the \ac{er} model from \cref{fig:er_example}.}
	\label{fig:db_example}
\end{figure}
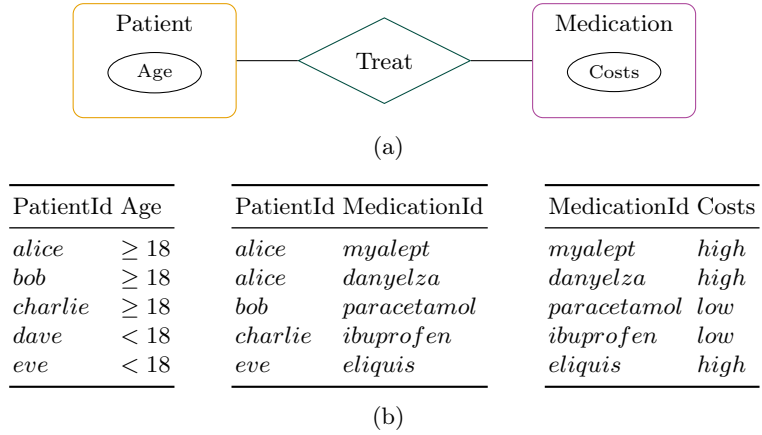
\begin{example} \label{ex:db_example}
	Take a look at a simple toy example database containing information on patients and the medications they take, as depicted in \cref{fig:db_example}.
	For simplicity, each patient only has a single attribute $Age$ which can either be $< 18$ or $\geq 18$ and each medication has a single attribute $Costs$ which can either be $low$ or $high$.
	Further, there is a relation $Treat$ connecting patients with the medications they take.
	The specific entries of the database are depicted in \cref{fig:db_table_example}, where we again keep the tables simple for illustrative purposes.

	A standard learning algorithm to obtain an \ac{fg} from the given database would include a single \ac{rv} for each of the attributes, e.g., there would be a single \ac{rv} for the attribute $Age$ instead of a separate node for the age of each patient.
	Analogously, there would also be a single \ac{rv} for the attribute $Costs$.
	When using a single \ac{rv} to model an attribute over all objects, however, we lose information about individual objects (here patients and medications) and their relationships between each other.
\end{example}
Therefore, we slightly adjust the learning procedure to include multiple \acp{rv} for the same attribute of different individual objects into the learned \ac{fg}.
However, we cannot simply include a \ac{rv} for each object per attribute because there would be no uncertainty in the resulting model.
To continue \cref{ex:db_example}, assume we add a \ac{rv} $Age$ for each patient to the learned \ac{fg}.
Then, the prior probability distribution for the \ac{rv} $Age$ of a specific patient would map the value found in the database for $Age$ to probability one and all other values to probability zero, e.g., the probability that $Age.alice < 18$ would be set to zero and the probability that $Age.alice \geq 18$ would be set to one.
Consequently, the \ac{fg} would not model any uncertainty at all.
To mitigate this issue, we propose to perform an initial clustering of entities to find clusters of indistinguishable objects (here patients and medications), which allows us to naturally define the domains of the \acp{lv} when transforming the learned \ac{fg} into an \ac{pfg}.
After the initial clustering, we then insert a \ac{rv} for each cluster per attribute into the \ac{fg}.
Performing an initial clustering of entities allows us to model uncertainty while keeping objects and relationships in the model.
\begin{example} \label{ex:db_clustering}
	Consider again the example database depicted in \cref{fig:db_example}.
	For the sake of the example, assume that the initial clustering returns two patient clusters $C_{p_1} = \{alice, eve\}$ as well as $C_{p_2} = \{bob, charlie, dave\}$ and two medication clusters $C_{m_1} = \{myalept, danyelza, eliquis\}$ and $C_{m_2} = \{paracetamol, ibuprofen\}$.
	Then, the resulting \ac{fg} contains the \acp{rv} $Age.p_1$, $Age.p_2$, $Costs.m_1$, and $Costs.m_2$, that is, there exists one \ac{rv} for each cluster per attribute.
	The probability that, e.g., $Age.p_1 < 18$ is then set to $0.5$ (as one out of two entries belonging to the cluster $C_{p_1}$ has $Age < 18$) and the probability that $Age.p_2 < 18$ is set to $0.33$ (as one out of three entries belonging to the cluster $C_{p_2}$ has $Age < 18$).
\end{example}
To perform the clustering of the entities, an arbitrary clustering algorithm can be applied.
In particular, it is also possible to apply privacy-preserving clustering algorithms, e.g., to ensure differential privacy guarantees.
After clustering the entities, the variable nodes of the \ac{fg} can be inserted for each clustered entity.

More specifically, the resulting \ac{fg} contains a \ac{rv} for every attribute $A$ in the database per cluster of entities occurring in $A$ (i.e., if $A$ is an attribute of an entity $E$, there is a \ac{rv} for $A$ for each cluster of $E$ and if $A$ is an attribute of a relationship $R$, there is a \ac{rv} for $R$ for each combination of clusters of the entities occurring in $R$).
To account not only for the attributes but also for the relationships, there is another \ac{rv} in the \ac{fg} for every relationship $R$ in the database per combination of clusters of entities occurring in $R$.

In general, the \acp{rv} specifying the attributes do not have to be Boolean.
The \acp{rv} for the relationships, however, are always Boolean as they indicate whether a relationship exists.
\begin{example}
	Given the database from \cref{fig:db_example} and the clusters $C_{p_1}$, $C_{p_2}$, $C_{m_1}$, and $C_{m_2}$ from \cref{ex:db_clustering}, the resulting \ac{fg} contains the \acp{rv} $Age.p_1$, $Age.p_2$, $Costs.m_1$, $Costs.m_2$, $Treat.p_1.m_1$, $Treat.p_1.m_2$, $Treat.p_2.m_1$, and $Treat.p_2.m_2$.
\end{example}
Note that in a standard learning algorithm for an \ac{fg}, there are typically no \acp{rv} for the relationships and, in addition to that, there are also not multiple \acp{rv} for the same attribute for different entities (or clusters of entities, respectively).
Therefore, an \ac{fg} learned by a standard learning algorithm loses information about the relationships of individual objects (or clusters thereof).

After having identified the variable nodes in the \ac{fg}, the next step is to learn the remaining graph structure, i.e., the factor nodes and the edges of the \ac{fg}.
The graph structure of the \ac{fg} can be identified using conditional independence tests on the given data~\cite[Chapter 20.7]{Koller2009a}.
In particular, two \acp{rv} $X$ and $Y$ are conditionally independent given a set of \acp{rv} $Z$ if $P(X, Y \mid Z) = P(X \mid Z) \cdot P(Y \mid Z)$.
Therefore, testing whether pairs of \acp{rv} are conditionally independent given a set of other \acp{rv} can be done by estimating the corresponding probabilities from the given data using statistical hypothesis tests.
The results of the conditional independence tests then yield the graph structure of the \ac{fg} as in an \ac{fg}, two \acp{rv} are conditionally independent if all paths between them are blocked by the conditioning set.

We remark that performing conditional independence tests on the individual tables does not work because the individual tables do not necessarily contain all \acp{rv} involved in the conditional independence test.
The conditional independence tests are hence carried out on a join of the individual tables in the given relational database to enable the estimation of the necessary probabilities.
\begin{example}
	Consider again the database given in \cref{fig:db_example}.
	When checking whether the \acp{rv} $Patient$ and $Medication$ are independent, we need a joined table which contains both $Patient$ and $Medication$.
\end{example}
As we add \acp{rv} for the relationships as well, we augment the full join with an additional column for each relationship, indicating which relationships are present in the database and which are not.
We also augment the full join with combinations of entities that do not occur in the relationships from the given database (similar to a cross join) such as the combination of patient $bob$ and medication $myalept$ in \cref{fig:db_example}, for which we add a row with value $false$ in the column $Treat$.
More details about the augmented full join are given in \cref{appendix:aug_full_join}.

After the entire graph structure of the \ac{fg} has been learned, the next step is to learn the potentials of the factors.
More specifically, at this point we know which factors (i.e., functions) are part of the model, but we do not know their input-output mappings yet.
To obtain the potentials of the factors, we count the occurrences for each specific assignment of the arguments of the factors in the augmented full join of the tables.
To count the occurrences for all possible assignments of factors' arguments, we also need information on combinations of entities that do not occur in the relationships from the given database, which is the reason we add this information in the augmented full join as described above.
The occurrences for each specific assignment of factors arguments are then counted as illustrated in the following example.
\begin{figure}[t]
	\centering
	\input{files/learn_example_01.tex}
	\caption{The learned \ac{fg} from the given database depicted in \cref{fig:db_example}.}
	\label{fig:learn_example_01}
\end{figure}
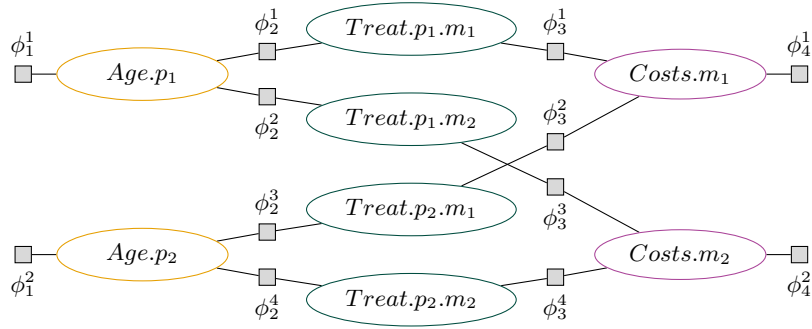
\begin{example}
	Take a look at the \ac{fg} shown in \cref{fig:learn_example_01}.
	The \ac{fg} results from the learning procedure described above applied to the database illustrated in \cref{fig:db_example}.
	The potentials for, e.g., $\phi_1^2(Age.p_2)$ are obtained by counting the occurrences of $Age.p_2 < 18$ and $Age.p_2 \geq 18$ in the augmented full join shown in \cref{fig:db_example_full_join} (\cref{appendix:aug_full_join}).
	Recall that $C_{p_2} = \{bob, charlie, dave\}$.
	Thus, in this particular example, it holds that $\phi_1^2(Age.p_2 < 18) = 5$ and $\phi_1^2(Age.p_2 \geq 18) = 10$.
\end{example}
Note that the absolute number of a potential value is not pivotal as the semantics of the \ac{fg} includes a normalisation constant.
Hence, the semantics would remain unchanged if we had counted the occurrences of $Age.p_2 < 18$ and $Age.p_2 \geq 18$ in the original table.
In the original table, we have one occurrence of $Age.p_2 < 18$ and two occurrences of $Age.p_2 \geq 18$ for the cluster $C_{p_2} = \{bob, charlie, dave\}$, i.e., the ratio is exactly the same as in the augmented full join (five occurrences and ten occurrences).
However, we use the augmented full join for counting as we cannot count occurrences of \acp{rv} appearing in different tables separately in factors with multiple arguments such as in $\phi_2^3(Age.p_2, Treat.p_2.m_1)$ because we have to incorporate the relationship between those tables.
For example, in $\phi_2^3(Age.p_2, Treat.p_2.m_1)$, we need to count the occurrences of $Age.p_2 < 18$ \emph{and} $Treat.p_2.m_1 = true$ and such a combination is not present in the original tables.

A summary of our proposed learning algorithm is provided in \cref{alg:learn_pfg}.
\begin{algorithm}[t]
	\SetKwInOut{Input}{Input}
	\SetKwInOut{Output}{Output}
	\caption{\learnpfg{}}
	\label{alg:learn_pfg}
	\Input{A relational database $\mathcal D = (\mathcal E, \mathcal R)$ with corresponding data samples.}
	\Output{A \ac{pfg} $G' = (\boldsymbol V, \boldsymbol E)$ representing the full joint probability distribution of the given database.}
	\BlankLine
	$G \gets$ Empty \ac{fg}\; \label{line:first_line}
	$F \gets$ Augmented full join over $\mathcal D$\;
	$C_{E_1}, \ldots, C_{E_{\ell}} \gets$ \textsc{ClusterEntities}($\mathcal D$)\;
	\ForEach{entity or relationship $B \in \mathcal E \cup \mathcal R$}{
		\tcp{Let $E_1, \ldots, E_j$ denote all entities occurring in $B$}
		\ForEach{combination of clusters $(c_1, \ldots, c_j) \in \times_{i=1}^{j} C_{E_i}$}{
			\ForEach{attribute $A \in \mathcal A(B)$}{
				Add a \ac{rv} $A.c_1.\cdots.c_j$ to $G$\;
			}
		}
	}
	\ForEach{relationship $R$}{
		\tcp{Let $E_1, \ldots, E_j$ denote all entities occurring in $R$}
		\ForEach{combination of clusters $(c_1, \ldots, c_j) \in \times_{i=1}^{j} C_{E_i}$}{
			Add a \ac{rv} $R.c_1.\cdots.c_j$ to $G$\;
		}
	}
	Add edges and factors to $G$ by running conditional independence tests on $F$\;
	\ForEach{factor $\phi(R_1, \ldots, R_k)$ in $G$}{
		\tcp{Let $c_1, \ldots, c_j$ denote all clusters occurring in $R_1, \ldots, R_k$}
		\ForEach{assignment $(r_1, \ldots, r_k) \in \times_{i=1}^{k} \range{R_i}$}{
			Set $\phi(R_1, \ldots, R_k)$ to the number of rows in $F$ belonging to clusters $c_1$ to $c_j$ that contain the values $r_1$ to $r_k$ in the columns of $R_1$ to $R_k$\;
		}
	} \label{line:last_for}
	$G' \gets$ Call \ac{acp} on $G$ with evidence $E = \emptyset$\; \label{line:call_acp}
	\Return{$G'$}
\end{algorithm}
So far, we discussed the steps to learn a propositional \ac{fg} from a given relational database (\cref{line:first_line} to \cref{line:last_for} in \cref{alg:learn_pfg}).
We next describe the procedure to transform the learned \ac{fg} into a \ac{pfg} (\cref{line:call_acp} in \cref{alg:learn_pfg}).

\subsection{Transforming the Factor Graph into a Parametric Factor Graph}
To obtain a \ac{pfg} from a given \ac{fg}, we need to find groups of identically behaving \acp{rv} and factors in the \ac{fg}.
Then, \acp{prv} with \acp{lv} represent such groups of indistinguishable \acp{rv} and \acp{pf} represent groups of identical factors.
Replacing indistinguishable \acp{rv} by \acp{prv} with \acp{lv} further abstracts from individuals and thus yields a promising foundation for privacy guarantees~\cite{Gehrke2024a}.
The \ac{acp} algorithm (which is a generalisation of the \acl{cp} algorithm~\cite{Ahmadi2013a,Kersting2009a}) is able to construct a \ac{pfg} from a given propositional \ac{fg}~\cite{Luttermann2024a}.
The idea behind \ac{acp} is to exploit symmetries in a propositional \ac{fg} and then group together symmetric subgraphs.
\Ac{acp} looks for symmetries based on potentials of factors, on ranges and evidence of \acp{rv}, as well as on the graph structure by passing around colours.
A formal description as well as an example run of the \ac{acp} algorithm can be found in \cref{appendix:acp_description}.
\Cref{fig:learn_example_02} shows the \ac{pfg} resulting from calling \ac{acp} on the \ac{fg} depicted in \cref{fig:learn_example_01} under the assumption that all potentials of the factors $\phi_i$, $i \in \{1,\ldots,4\}$, are considered identical.
Note that the assumption of identical factors is just for the sake of the example as in general, not all potentials are identical (and hence, not all of the factors $\phi_i$ are grouped into a single group).

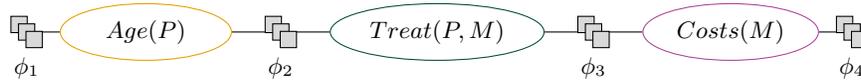
\begin{figure}[t]
	\centering
	\input{files/learn_example_02.tex}
	\caption{The \ac{pfg} resulting from calling \ac{acp} on the \ac{fg} depicted in \cref{fig:learn_example_01} assuming for the sake of the example that all clusters behave identically. Note that in our specific example from \cref{ex:db_example}, not all clusters behave identically and thus, not all \acp{rv} representing the same attribute or relationship are grouped.}
	\label{fig:learn_example_02}
\end{figure}

We remark that in its original form, \ac{acp} requires potentials of factors to identically match in order to group factors together.
When learning an \ac{fg} from data, however, there might be small deviations in the potentials, that is, counting the occurrences of a specific combination of values might differ for two similarly behaving clusters, e.g., by one.
Therefore, we adapt the condition for two factors to be considered identical by allowing for a difference in their potentials depending on a user-defined parameter $\varepsilon$.
Two factors $\phi$ and $\phi'$ are now considered identical (and thus are assigned the same colour in \ac{acp}) if for every assignment of their arguments the corresponding potentials, say $\varphi$ and $\varphi'$, differ by at most $\varphi \cdot \epsilon$ (or $\varphi' \cdot \epsilon$, respectively).
Factors with slightly deviating potentials that are grouped are then all assigned the mean of the potentials in the group.

\subsection{Sampling from the Parametric Factor Graph}
Every \ac{pfg} compactly encodes a full joint probability distribution from which we can draw new samples.
Note that, in contrast to previous sampling approaches (such as, e.g., \cite{Venugopal2014a}) that apply sampling for (approximate) query answering, we aim to generate new data samples that comply with the given \ac{er} model (that is, we wish to draw new data samples that contain a value for every attribute and every relationship in the \ac{er} model).
As the \ac{pfg} is inherently encoding a relational structure, we are able to synthesise relational data.
More specifically, the semantics of a \ac{pfg} is defined on a ground level, that is, sampling from the underlying probability distribution yields new data samples for multiple clustered entities.
As the ground \ac{fg} consists of \acp{rv} for various clustered entities, we draw samples for multiple clusters at the same time.
\begin{example}
	Assume we want to sample the \ac{pfg} which yields the \ac{fg} shown in \cref{fig:learn_example_01} when grounding the model.
	Sampling thus yields a value for every \ac{rv} in the \ac{fg}, thereby allowing to synthesise new objects and relationships between them following the full joint probability distribution encoded by the model.
	An exemplary data sample could look like this: $Age.p_1 < 18$, $Age.p_2 \geq 18$, $Treat.p_1.m_1 = false$, $Treat.p_1.m_2 = false$, $Treat.p_2.m_1 = true$, $Treat.p_2.m_2 = false$, $Costs.m_1 = low$, $Costs.m_2 = high$ (values are chosen arbitrarily for the sake of the example).
\end{example}
The generated synthetic data samples might be used for arbitrary applications and thus, it is necessary to define application-specific quality criteria to assess the quality of the generated data.

\section{Conclusion}
We introduce a fully-fledged pipeline to deploy probabilistic relational models, in particular \acp{pfg}, for the generation of synthetic relational (i.e., multi-table) data.
To construct a \ac{pfg} from a given relational database, we propose a learning algorithm that learns both the graph structure as well as the parameters of a \ac{pfg} from the relational database.
We further elaborate on how the learned \ac{pfg} can be applied to generate new samples of synthetic relational data.
By ensuring certain privacy guarantees (e.g., differential privacy) during the construction process of the \ac{pfg}, \ac{pfg} provide a promising model to generate synthetic relational data in a privacy-preserving manner such that generated synthetic data can be publicly shared without leaking sensitive data of individuals.

There are three primary directions for future work.
First, privacy guarantees for \ac{pfg} learning need to be theoretically investigated.
Second, the scalability of our proposed architecture should be assessed and improved to allow an efficient handling of large-scale relational databases, and finally, the practicality of the entire framework has to be tested empirically on real-world data sets.

\section*{Acknowledgements}
This work is funded by the BMBF project AnoMed 16KISA057.
This preprint has not undergone peer review or any post-submission improvements or corrections.
The Version of Record of this contribution is published in \emph{Lecture Notes in Computer Science, Volume 14992}, and is available online at \url{https://doi.org/10.1007/978-3-031-70893-0_13}.

\bibliographystyle{splncs04}
\bibliography{references}

\clearpage
\appendix

\section{Augmented Full Join} \label{appendix:aug_full_join}
We call the full join of the tables, where an additional column for each relationship is added and missing relationships are encoded by an additional row containing the value $false$ in the corresponding relationship column, the \emph{augmented full join}.
The augmented full join can therefore be thought of as a cross join with an additional column for each relationship, which contains a Boolean value indicating which relationships are present in the database.
For example, the augmented full join of the tables given in the example from \cref{fig:db_example} is illustrated in \cref{fig:db_example_full_join}.
In this particular example, the augmented full join contains the additional column $Treat$, which contains the value $true$ if the relationship of a given combination of $PatientId$ and $MedicationId$ in a specific row actually exists.
Otherwise, the column $Treat$ contains the value $false$.
\begin{figure}
	\centering
	\input{files/db_example_full_join.tex}
	\caption{Full join of the tables from \cref{fig:db_table_example}, where missing entries for the relationship $Treat$ have been added by setting the value of the column $Treat$ to $false$.}
	\label{fig:db_example_full_join}
\end{figure}

\section{Formal Description of the Advanced Colour Passing Algorithm} \label{appendix:acp_description}
The \ac{acp} algorithm~\cite{Luttermann2024a} builds on the \acl{cp} algorithm~\cite{Ahmadi2013a,Kersting2009a} and solves the problem of constructing a \ac{pfg} from a given \ac{fg}.
\Cref{alg:acp} provides a formal description of the \ac{acp} algorithm.
\begin{algorithm}
	\SetKwInOut{Input}{Input}
	\SetKwInOut{Output}{Output}
	\caption{Advanced Colour Passing (as introduced in \cite{Luttermann2024a})}
	\label{alg:acp}
	\Input{An \ac{fg} $G$ with \acp{rv} $\boldsymbol R = \{R_1, \ldots, R_n\}$, and factors $\boldsymbol \Phi = \{\phi_1, \ldots, \phi_m\}$, as well as a set of evidence $\boldsymbol E = \{R_1 = r_1, \ldots, R_k = r_k\}$.}
	\Output{A lifted representation $G'$ in form of a \ac{pfg} with equivalent semantics to $G$.}
	\BlankLine
	Assign each $R_i$ a colour according to $\mathcal R(R_i)$ and $\boldsymbol E$\;
	Assign each $\phi_i$ a colour according to order-independent potentials and rearrange arguments accordingly\;
	\Repeat{grouping does not change}{
		\ForEach{factor $\phi \in \boldsymbol \Phi$}{
			$signature_{\phi} \gets [\,]$\;
			\ForEach{\ac{rv} $R \in neighbours(G, \phi)$}{
				\tcp{In order of appearance in $\phi$}
				$append(signature_{\phi}, R.colour)$\;
			}
			$append(signature_{\phi}, \phi.colour)$\;
		}
		Group together all $\phi$s with the same signature\;
		Assign each such cluster a unique colour\;
		Set $\phi.colour$ correspondingly for all $\phi$s\;
		\ForEach{\ac{rv} $R \in \boldsymbol R$}{
			$signature_{R} \gets [\,]$\;
			\ForEach{factor $\phi \in neighbours(G, R)$}{
				\uIf{$\phi$ is commutative w.r.t.\ $\boldsymbol S$ and $R \in \boldsymbol S$}{
					$append(signature_{R}, (\phi.colour, 0))$\;
				}
				\Else{
					$append(signature_{R}, (\phi.colour, p(R, \phi)))$\;
				}
			}
			Sort $signature_{R}$ according to colour\;
			$append(signature_{R}, R.colour)$\;
		}
		Group together all $R$s with the same signature\;
		Assign each such cluster a unique colour\;
		Set $R.colour$ correspondingly for all $R$s\;
	}
	$G' \gets$ construct \acs{pfg} from groupings\;
\end{algorithm}
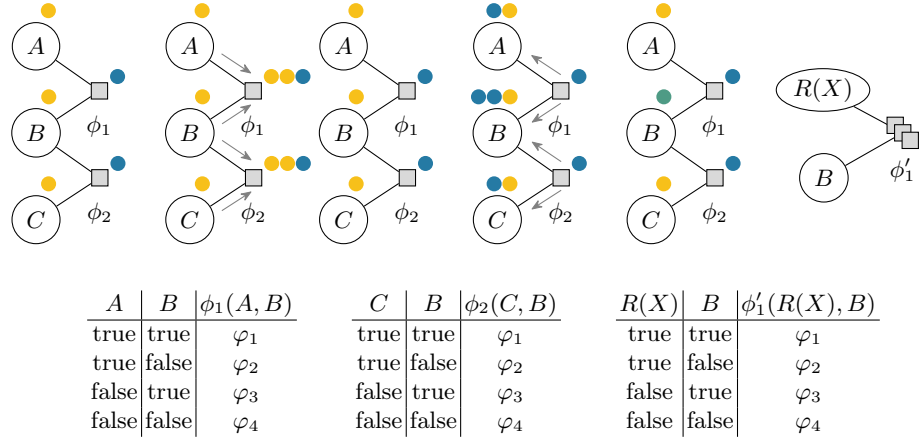
\begin{figure}[t]
	\centering
	\input{files/acp_example.tex}
		\caption{A visualisation of the steps undertaken by the \ac{acp} algorithm on an input \ac{fg} with only Boolean \acp{rv} and no evidence (left). Colours are first passed from variable nodes to factor nodes, followed by a recolouring, and then passed back from factor nodes to variable nodes, again followed by a recolouring. The procedure is iterated until convergence and the resulting \ac{pfg} is depicted on the right. This figure is reprinted from~\protect\cite{Luttermann2024a}.}
	\label{fig:acp_example}
\end{figure}

\Cref{fig:acp_example} illustrates the \ac{acp} algorithm on an example \ac{fg}~\cite{Ahmadi2013a}.
In this example, $A$, $B$, and $C$ are Boolean \acp{rv} with no evidence and thus, they all receive the same colour (e.g., $\mathrm{yellow}$).
As the potentials of $\phi_1$ and $\phi_2$ are identical, $\phi_1$ and $\phi_2$ are assigned the same colour as well (e.g., $\mathrm{blue}$)\footnote{The \ac{deft} algorithm~\cite{Luttermann2024d} can be applied to efficiently detect factors that encode identical potentials regardless of their argument orders and to rearrange the factors' arguments accordingly.}.
The colour passing then starts from variable nodes to factor nodes, that is, $A$ and $B$ send their colour ($\mathrm{yellow}$) to $\phi_1$ and $B$ and $C$ send their colour ($\mathrm{yellow}$) to $\phi_2$.
$\phi_1$ and $\phi_2$ are then recoloured according to the colours they received from their neighbours to reduce the communication overhead.
Since $\phi_1$ and $\phi_2$ received identical colours (two times the colour $\mathrm{yellow}$), they are assigned the same colour during recolouring.
Afterwards, the colours are passed from factor nodes to variable nodes and this time not only the colours but also the position of the \acp{rv} in the argument list of the corresponding factor are shared.
Consequently, $\phi_1$ sends a tuple $(\mathrm{blue}, 1)$ to $A$ and a tuple $(\mathrm{blue}, 2)$ to $B$, and $\phi_2$ sends a tuple $(\mathrm{blue}, 2)$ to $B$ and a tuple $(\mathrm{blue}, 1)$ to $C$ (positions are not shown in \cref{fig:acp_example}).
Since $A$ and $C$ are both at position one in the argument list of their respective neighbouring factor, they receive identical messages and are recoloured with the same colour.
$B$ is assigned a different colour during recolouring than $A$ and $C$ because $B$ received different messages than $A$ and $C$.
The groupings do not change in further iterations and hence the algorithm terminates.
The output is the \ac{pfg} shown on the right in \cref{fig:acp_example}, where both $A$ and $C$ as well as $\phi_1$ and $\phi_2$ are grouped.

\end{document}

%% file: defs.tex
\acrodef{acp}[ACP]{advanced colour passing}
\acrodef{bn}[BN]{Bayesian network}
\acrodef{cp}[CP]{colour passing}
\acrodef{crv}[CRV]{counting randvar}
\acrodef{decor}[DECOR]{detection of commutative factors}
\acrodef{deft}[DEFT]{detection of exchangeable factors}
\acrodef{er}[ER]{entity-relationship}
\acrodef{fg}[FG]{factor graph}
\acrodef{ljt}[LJT]{lifted junction tree}
\acrodef{lv}[logvar]{logical variable}
\acrodef{lve}[LVE]{lifted variable elimination}
\acrodef{mln}[MLN]{Markov logic network}
\acrodef{mn}[MN]{Markov network}
\acrodef{pcrv}[PCRV]{parameterised CRV}
\acrodef{pf}[parfactor]{parametric factor}
\acrodef{pfg}[PFG]{parametric factor graph}
\acrodef{prv}[PRV]{parameterised randvar}
\acrodef{rv}[randvar]{random variable}
\acrodef{ve}[VE]{variable elimination}
\acrodef{wl}[WL]{Weisfeiler-Leman}

\crefname{algocf}{Alg.}{Algs.}
\Crefname{algocf}{Algorithm}{Algorithms}
\crefname{algorithm}{Alg.}{Algs.}
\Crefname{algorithm}{Algorithm}{Algorithms}
\crefname{corollary}{Cor.}{Cors.}
\Crefname{corollary}{Corollary}{Corollaries}
\crefname{definition}{Def.}{Defs.}
\Crefname{definition}{Definition}{Definitions}
\crefname{example}{Ex.}{Exs.}
\Crefname{example}{Example}{Examples}
\crefname{proposition}{Prop.}{Props.}
\Crefname{proposition}{Proposition}{Propositions}
\crefname{section}{Sec.}{Secs.}
\Crefname{section}{Section}{Sections}
\crefname{theorem}{Thm.}{Thms.}
\Crefname{theorem}{Theorem}{Theorems}

\newcommand{\domain}[1]{\ensuremath{\mathrm{dom}(#1)}}
\newcommand{\learnpfg}{\textsc{LearnPFG}}
\newcommand{\range}[1]{\ensuremath{\mathrm{range}(#1)}}

\definecolor{myyellow}{RGB}{247,192,26}
\definecolor{myblue}{RGB}{37,122,164}
\definecolor{mygreen}{RGB}{78,155,133}
\definecolor{mypurple}{RGB}{86,51,94}

\definecolor{newblue}{RGB}{50,113,173}
\definecolor{newred}{RGB}{222,32,36}
\definecolor{newgreen}{RGB}{70,165,69}
\definecolor{newpurple}{RGB}{140,69,152}

\definecolor{cborange}{RGB}{230,159,0}
\definecolor{cbblue}{RGB}{30,136,229}
\definecolor{cbbluedark}{RGB}{46,37,133}
\definecolor{cbpurple}{RGB}{170,68,153}
\definecolor{cbgreen}{RGB}{0,77,64}
\definecolor{cbgreenlight}{RGB}{93,168,153}
\definecolor{cbbrown}{RGB}{126,41,84}

\pgfdeclarelayer{bg}
\pgfsetlayers{bg,main}

\tikzset{
	rv/.style={draw, ellipse},
	pf/.style={draw, rectangle, fill = gray!30},
	arc/.style = {->, >={[round,sep]Stealth}},
}

\newcommand\factorat[4]{
	\node[pf, label={#2:{#3}}](#4) at (#1) {};
}

\newcommand\factor[6]{
	\node[pf, #1=#3 of #2, label={#4:{#5}}](#6) {};
}

\newcommand\nodecolorshift[5]{
	\node[circle, fill=#1, above right=0.1cm of #2, inner sep=0pt, minimum size=2mm, xshift=#4, yshift=#5](#3) {};
}

\newcommand\factorcolor[3]{
	\node[circle, fill=#1, above right=0cm and 0.05cm of #2, inner sep=0pt, minimum size=2mm](#3) {};
}
\newcommand\factorcolorshift[4]{
	\node[circle, fill=#1, above right=0cm and 0.05cm of #2, inner sep=0pt, minimum size=2mm, xshift=#4](#3) {};
}

\newcommand\pfs[8]{
	\node[pf, #1=#3 of #2, xshift=-1mm, yshift=1mm](#6) {};
	\node[pf, #1=#3 of #2, label={[label distance=1mm]#4:{#5}}](#7) {};
	\node[pf, #1=#3 of #2, xshift=1mm, yshift=-1mm](#8) {};
}

\newcommand\pfsat[6]{
	\node[pf, xshift=-1mm, yshift=1mm](#4) at (#1) {};
	\node[pf, label={[label distance=1mm]#2:{#3}}](#5) at (#1) {};
	\node[pf, xshift=1mm, yshift=-1mm](#6) at (#1) {};
}

%% file: files/example_fg.tex
\begin{tikzpicture}[
	rv/.append style={fill=white}
]
	\node[rv] (E) {$Epid$};

	\factor{above left}{E}{0.2cm and 0.3cm}{[label distance=-1mm]90}{$\phi_0$}{F0}

	\factor{below left}{E}{-0.1cm and 2cm}{[label distance=-2mm]315}{$\phi_1$}{F1_1}
	\factor{below right}{E}{-0.1cm and 2cm}{[label distance=-2mm]215}{$\phi_1$}{F1_2}
	\factor{above left}{E}{-0.1cm and 2cm}{[label distance=-2mm]45}{$\phi_1$}{F1_3}
	\factor{above right}{E}{-0.1cm and 2cm}{[label distance=-2mm]135}{$\phi_1$}{F1_4}

	\node[rv, draw=cbblue, below = 0.6cm of F1_1] (SickA) {$Sick.alice$};
	\node[rv, draw=cbblue, below = 0.6cm of F1_2] (SickB) {$Sick.bob$};
	\node[rv, draw=cbblue, above = 0.6cm of F1_3] (SickD) {$Sick.dave$};
	\node[rv, draw=cbblue, above = 0.6cm of F1_4] (SickE) {$Sick.eve$};

	\factor{right}{SickA}{0.5cm}{[label distance=-1mm]270}{$\phi_2$}{F2_1}
	\factor{below right}{F2_1}{0.5cm and 0.2cm}{[label distance=-1mm]270}{$\phi_2$}{F2_2}

	\factor{left}{SickB}{0.5cm}{[label distance=-1mm]270}{$\phi_2$}{F2_3}
	\factor{below left}{F2_3}{0.5cm and 0.2cm}{[label distance=-1mm]270}{$\phi_2$}{F2_4}

	\factor{right}{SickD}{0.5cm}{[label distance=-1mm]90}{$\phi_2$}{F2_5}
	\factor{above right}{F2_5}{0.5cm and 0.2cm}{[label distance=-1mm]90}{$\phi_2$}{F2_6}

	\factor{left}{SickE}{0.5cm}{[label distance=-1mm]90}{$\phi_2$}{F2_7}
	\factor{above left}{F2_7}{0.5cm and 0.2cm}{[label distance=-1mm]90}{$\phi_2$}{F2_8}

	\node[rv, draw=cborange, below left = 0.0cm and 0.5cm of F1_1] (TravelA) {$Travel.alice$};
	\node[rv, draw=cborange, below right = 0.0cm and 0.5cm of F1_2] (TravelB) {$Travel.bob$};
	\node[rv, draw=cborange, above left = 0.0cm and 0.5cm of F1_3] (TravelD) {$Travel.dave$};
	\node[rv, draw=cborange, above right = 0.0cm and 0.5cm of F1_4] (TravelE) {$Travel.eve$};
	\node[rv, draw=cbgreen, below left = 0.2cm and -0.8cm of SickA] (TreatAM1) {$Treat.alice.m_1$};
	\node[rv, draw=cbgreen, below right = 0.2cm and -0.45cm of TreatAM1] (TreatAM2) {$Treat.alice.m_2$};
	\node[rv, draw=cbgreen, below right = 0.2cm and -0.8cm of SickB] (TreatBM1) {$Treat.bob.m_1$};
	\node[rv, draw=cbgreen, below left = 0.2cm and -0.45cm of TreatBM1] (TreatBM2) {$Treat.bob.m_2$};
	\node[rv, draw=cbgreen, above left = 0.2cm and -0.8cm of SickD] (TreatDM1) {$Treat.dave.m_1$};
	\node[rv, draw=cbgreen, above right = 0.2cm and -0.45cm of TreatDM1] (TreatDM2) {$Treat.dave.m_2$};
	\node[rv, draw=cbgreen, above right = 0.2cm and -0.8cm of SickE] (TreatEM1) {$Treat.eve.m_1$};
	\node[rv, draw=cbgreen, above left = 0.2cm and -0.45cm of TreatEM1] (TreatEM2) {$Treat.eve.m_2$};

	\factor{left}{TravelA}{0.3cm}{[label distance=-1mm]270}{$\phi_3$}{F3_1}
	\factor{right}{TravelB}{0.3cm}{[label distance=-1mm]270}{$\phi_3$}{F3_2}
	\factor{left}{TravelD}{0.3cm}{[label distance=-1mm]90}{$\phi_3$}{F3_3}
	\factor{right}{TravelE}{0.3cm}{[label distance=-1mm]90}{$\phi_3$}{F3_4}

	\begin{pgfonlayer}{bg}
		\draw (E) -- (F0);
		\draw (E) -- (F1_1);
		\draw (E) -- (F2_1);
		\draw (E) -- (F2_2);
		\draw (E) -- (F1_2);
		\draw (E) -- (F2_3);
		\draw (E) -- (F1_3);
		\draw (E) -- (F2_4);
		\draw (E) -- (F1_4);
		\draw (E) -- (F2_5);
		\draw (E) -- (F2_6);
		\draw (E) -- (F2_7);
		\draw (E) -- (F2_8);
		\draw (SickA) -- (F1_1);
		\draw (SickA) -- (F2_1);
		\draw (SickA) -- (F2_2);
		\draw (TravelA) -- (F1_1);
		\draw (TreatAM1.east) -- (F2_1);
		\draw (TreatAM2) -- (F2_2);
		\draw (SickB) -- (F1_2);
		\draw (SickB) -- (F2_3);
		\draw (SickB) -- (F2_4);
		\draw (TravelB) -- (F1_2);
		\draw (TreatBM1.west) -- (F2_3);
		\draw (TreatBM2) -- (F2_4);
		\draw (SickD) -- (F1_3);
		\draw (SickD) -- (F2_5);
		\draw (SickD) -- (F2_6);
		\draw (TravelD) -- (F1_3);
		\draw (TreatDM1.east) -- (F2_5);
		\draw (TreatDM2) -- (F2_6);
		\draw (SickE) -- (F1_4);
		\draw (SickE) -- (F2_7);
		\draw (SickE) -- (F2_8);
		\draw (TravelE) -- (F1_4);
		\draw (TreatEM1.west) -- (F2_7);
		\draw (TreatEM2) -- (F2_8);
		\draw (TravelA) -- (F3_1);
		\draw (TravelB) -- (F3_2);
		\draw (TravelD) -- (F3_3);
		\draw (TravelE) -- (F3_4);
	\end{pgfonlayer}
\end{tikzpicture}

%% file: files/example_pfg.tex
\begin{tikzpicture}
	\node[rv] (E) {$Epid$};
	\node[rv, draw=cbblue, below = 0.3cm of E] (S) {$Sick(P)$};
	\node[rv, draw=cborange, left = 0.3cm of S] (Travel) {$Travel(P)$};
	\node[rv, draw=cbgreen, right = 0.3cm of S] (Treat) {$Treat(P,M)$};
	\factor{above}{E}{0.15cm}{90}{$\phi_0$}{G0}
	\pfs{above}{Travel}{0.55cm}{90}{$\phi_1$}{G1a}{G1}{G1b}
	\pfs{above}{Treat}{0.55cm}{90}{$\phi_2$}{G2a}{G2}{G2b}
	\pfs{left}{Travel}{0.25cm}{180}{$\phi_3$}{G3a}{G3}{G3b}

	\begin{pgfonlayer}{bg}
		\draw (E) -- (G0);
		\draw (E) -- (G1);
		\draw (E) -- (G2);
		\draw (S) -- (G1);
		\draw (S) -- (G2);
		\draw (Travel) -- (G1);
		\draw (Treat) -- (G2);
		\draw (Travel) -- (G3);
	\end{pgfonlayer}
\end{tikzpicture}

%% file: files/architecture_overview.tex
\begin{center}
	\begin{minipage}{0.07\textwidth}
		\centering
		\scalebox{0.5}{
			\tikz{
				\node[cylinder,draw=black,fill=gray!20,shape border rotate=90,minimum height=3.5em,minimum width=3.0em] (db) {};
			}
		}
	\end{minipage}
	\tikz{\draw[arc] (0,0) -- (0.35,0);}
	\hskip 0.05cm
	\begin{minipage}{0.55\textwidth}
		\centering
		\scalebox{0.6}{\input{files/learn_example_01.tex}}
	\end{minipage}
	\tikz{\draw[arc] (0,0) -- (0.35,0);}
	\hskip 0.05cm
	\begin{minipage}{0.16\textwidth}
		\centering
		\scalebox{0.6}{\input{files/learn_example_02_vertical.tex}}
	\end{minipage}
	$+$
	\begin{minipage}{0.07\textwidth}
		\centering
		\scalebox{0.5}{
			\tikz{
				\node[cylinder,draw=black,fill=gray!20,shape border rotate=90,minimum height=3.5em,minimum width=3.0em] (db) {};
			}
		}
	\end{minipage}
\end{center}

%% file: files/learn_example_01.tex
\begin{tikzpicture}[rv/.append style={fill=white,minimum width=7.0em}]
	\node[rv, draw=cbgreen] (T11) {$Treat.p_1.m_1$};
	\node[rv, draw=cbgreen, below = 1.5em of T11] (T12) {$Treat.p_1.m_2$};
	\node[rv, draw=cbgreen, below = 1.5em of T12] (T21) {$Treat.p_2.m_1$};
	\node[rv, draw=cbgreen, below = 1.5em of T21] (T22) {$Treat.p_2.m_2$};

	\node[rv, draw=cborange, xshift = -11.0em] at ($(T11)!0.5!(T12)$) (A1) {$Age.p_1$};
	\node[rv, draw=cborange, xshift = -11.0em] at ($(T21)!0.5!(T22)$) (A2) {$Age.p_2$};
	
	\node[rv, draw=cbpurple, xshift = 11.0em] at ($(T11)!0.5!(T12)$) (C1) {$Costs.m_1$};
	\node[rv, draw=cbpurple, xshift = 11.0em] at ($(T21)!0.5!(T22)$) (C2) {$Costs.m_2$};

	\factor{left}{A1}{1.0em}{90}{$\phi_{1}^{1}$}{f11}
	\factor{left}{A2}{1.0em}{270}{$\phi_{1}^{2}$}{f12}

	\factor{right}{C1}{1.0em}{90}{$\phi_{4}^{1}$}{f41}
	\factor{right}{C2}{1.0em}{270}{$\phi_{4}^{2}$}{f42}

	\factorat{$(A1.east)!0.5!(T11.west)$}{90}{$\phi_{2}^{1}$}{f21}
	\factorat{$(A1.east)!0.5!(T12.west)$}{270}{$\phi_{2}^{2}$}{f22}
	\factorat{$(A2.east)!0.5!(T21.west)$}{90}{$\phi_{2}^{3}$}{f23}
	\factorat{$(A2.east)!0.5!(T22.west)$}{270}{$\phi_{2}^{4}$}{f24}

	\factorat{$(C1.west)!0.5!(T11.east)$}{90}{$\phi_{3}^{1}$}{f31}
	\factorat{$(C1.west)!0.5!(T21.east)$}{90}{$\phi_{3}^{2}$}{f32}
	\factorat{$(C2.west)!0.5!(T12.east)$}{270}{$\phi_{3}^{3}$}{f33}
	\factorat{$(C2.west)!0.5!(T22.east)$}{270}{$\phi_{3}^{4}$}{f34}

	\begin{pgfonlayer}{bg}
		\draw (A1) -- (f11);
		\draw (A2) -- (f12);
		\draw (C1) -- (f41);
		\draw (C2) -- (f42);
		\draw (A1) -- (f21);
		\draw (A1) -- (f22);
		\draw (f21) -- (T11);
		\draw (f22) -- (T12);
		\draw (A2) -- (f23);
		\draw (A2) -- (f24);
		\draw (f23) -- (T21);
		\draw (f24) -- (T22);
		\draw (C1) -- (f31);
		\draw (C1) -- (f32);
		\draw (f31) -- (T11);
		\draw (f32) -- (T21);
		\draw (C2) -- (f33);
		\draw (C2) -- (f34);
		\draw (f33) -- (T12);
		\draw (f34) -- (T22);
	\end{pgfonlayer}
\end{tikzpicture}

%% file: files/learn_example_02_vertical.tex
\begin{tikzpicture}[rv/.append style={fill=white,minimum width=7.0em}]
	\node[rv, draw=cbgreen] (T) {$Treat(P,M)$};
	\node[rv, draw=cborange, above = 2.5em of T] (A) {$Age(P)$};
	\node[rv, draw=cbpurple, below = 2.5em of T] (C) {$Costs(M)$};

	\pfs{above}{A}{1.0em}{180}{$\phi_1$}{G1a}{G1}{G1b}
	\pfs{below}{C}{1.0em}{180}{$\phi_4$}{G4a}{G4}{G4b}

	\pfsat{$(A.south)!0.5!(T.north)$}{180}{$\phi_2$}{G2a}{G2}{G2b}
	\pfsat{$(C.north)!0.5!(T.south)$}{180}{$\phi_3$}{G3a}{G3}{G3b}

	\begin{pgfonlayer}{bg}
		\draw (A) -- (G1);
		\draw (A) -- (G2);
		\draw (G2) -- (T);
		\draw (C) -- (G4);
		\draw (C) -- (G3);
		\draw (G3) -- (T);
	\end{pgfonlayer}
\end{tikzpicture}

%% file: files/er_example.tex
\begin{tikzpicture}
	\node[draw=cborange,rectangle,rounded corners,text width=6.0em,text height=4.0em] (1) {};
	\node[above=-1.5em of 1] (1l) {\footnotesize Patient};
	\node[draw,ellipse,below=0.5em of 1l,align=center,text width=2.2em,minimum height=1.6em,inner sep=0.25em] (1a1) {\scriptsize Age};

	\node[draw=cbgreen,diamond,aspect=2,align=center,text width=4.5em,right=2.5em of 1,inner sep=0.2em] (2) {\footnotesize Treat};

	\node[draw=cbpurple,rectangle,rounded corners,text width=6.0em,text height=4.0em,right=2.5em of 2] (3) {};
	\node[above=-1.5em of 3] (3l) {\footnotesize Medication};
	\node[draw,ellipse,below=0.5em of 3l,align=center,text width=2.2em,minimum height=1.6em,inner sep=0.25em] (3a1) {\scriptsize Costs};

	\draw (1.east) -- (2.west);
	\draw (2.east) -- (3.west);
\end{tikzpicture}

%% file: files/db_example.tex
\vskip 0.75em
\begin{tabular}{ll}
	\toprule
	PatientId & Age       \\ \midrule
	$alice$   & $\geq 18$ \\
	$bob$     & $\geq 18$ \\
	$charlie$ & $\geq 18$ \\
	$dave$    & $< 18$    \\
	$eve$     & $< 18$    \\ \bottomrule
\end{tabular}
\qquad
\begin{tabular}{ll}
	\toprule
	PatientId & MedicationId  \\ \midrule
	$alice$   & $myalept$     \\
	$alice$   & $danyelza$    \\
	$bob$     & $paracetamol$ \\
	$charlie$ & $ibuprofen$   \\
	$eve$     & $eliquis$     \\ \bottomrule
\end{tabular}
\qquad
\begin{tabular}{ll}
	\toprule
	MedicationId & Costs  \\ \midrule
	$myalept$ & $high$    \\
	$danyelza$ & $high$   \\
	$paracetamol$ & $low$ \\
	$ibuprofen$ & $low$   \\
	$eliquis$ & $high$    \\ \bottomrule
\end{tabular}

%% file: files/learn_example_02.tex
\begin{tikzpicture}[rv/.append style={fill=white,minimum width=7.0em}]
	\node[rv, draw=cbgreen] (T) {$Treat(P,M)$};
	\node[rv, draw=cborange, left = 4.0em of T] (A) {$Age(P)$};
	\node[rv, draw=cbpurple, right = 4.0em of T] (C) {$Costs(M)$};

	\pfs{left}{A}{1.0em}{270}{$\phi_1$}{G1a}{G1}{G1b}
	\pfs{right}{C}{1.0em}{270}{$\phi_4$}{G4a}{G4}{G4b}

	\pfsat{$(A.east)!0.5!(T.west)$}{270}{$\phi_2$}{G2a}{G2}{G2b}
	\pfsat{$(C.west)!0.5!(T.east)$}{270}{$\phi_3$}{G3a}{G3}{G3b}

	\begin{pgfonlayer}{bg}
		\draw (A) -- (G1);
		\draw (A) -- (G2);
		\draw (G2) -- (T);
		\draw (C) -- (G4);
		\draw (C) -- (G3);
		\draw (G3) -- (T);
	\end{pgfonlayer}
\end{tikzpicture}

%% file: files/db_example_full_join.tex
\begin{tabular}{lllll}
	\toprule
	PatientId & Age       & MedicationId  & Treat   & Costs  \\ \midrule
	$alice$   & $\geq 18$ & $myalept$     & $true$  & $high$ \\
	$alice$   & $\geq 18$ & $danyelza$    & $true$  & $high$ \\
	$alice$   & $\geq 18$ & $paracetamol$ & $false$ & $low$  \\
	$alice$   & $\geq 18$ & $ibuprofen$   & $false$ & $low$  \\
	$alice$   & $\geq 18$ & $eliquis$     & $false$ & $high$ \\
	$bob$     & $\geq 18$ & $myalept$     & $false$ & $high$ \\
	$bob$     & $\geq 18$ & $danyelza$    & $false$ & $high$ \\
	$bob$     & $\geq 18$ & $paracetamol$ & $true$  & $low$  \\
	$bob$     & $\geq 18$ & $ibuprofen$   & $false$ & $low$  \\
	$bob$     & $\geq 18$ & $eliquis$     & $false$ & $high$ \\
	$charlie$ & $\geq 18$ & $myalept$     & $false$ & $high$ \\
	$charlie$ & $\geq 18$ & $danyelza$    & $false$ & $high$ \\
	$charlie$ & $\geq 18$ & $paracetamol$ & $false$ & $low$  \\
	$charlie$ & $\geq 18$ & $ibuprofen$   & $true$  & $low$  \\
	$charlie$ & $\geq 18$ & $eliquis$     & $false$ & $high$ \\
	$dave$    & $< 18$    & $myalept$     & $false$ & $high$ \\
	$dave$    & $< 18$    & $danyelza$    & $false$ & $high$ \\
	$dave$    & $< 18$    & $paracetamol$ & $false$ & $low$  \\
	$dave$    & $< 18$    & $ibuprofen$   & $false$ & $low$  \\
	$dave$    & $< 18$    & $eliquis$     & $false$ & $high$ \\
	$eve$     & $< 18$    & $myalept$     & $false$ & $high$ \\
	$eve$     & $< 18$    & $danyelza$    & $false$ & $high$ \\
	$eve$     & $< 18$    & $paracetamol$ & $false$ & $low$  \\
	$eve$     & $< 18$    & $ibuprofen$   & $false$ & $low$  \\
	$eve$     & $< 18$    & $eliquis$     & $true$  & $high$ \\ \bottomrule
\end{tabular}

%% file: files/acp_example.tex
\begin{tikzpicture}[label distance=1mm]
	\node[circle, draw] (A) {$A$};
	\node[circle, draw] (B) [below = 0.5cm of A] {$B$};
	\node[circle, draw] (C) [below = 0.5cm of B] {$C$};
	\factor{below right}{A}{0.25cm and 0.5cm}{270}{$\phi_1$}{f1}
	\factor{below right}{B}{0.25cm and 0.5cm}{270}{$\phi_2$}{f2}

	\nodecolorshift{myyellow}{A}{Acol}{-2.1mm}{1mm}
	\nodecolorshift{myyellow}{B}{Bcol}{-2.1mm}{1mm}
	\nodecolorshift{myyellow}{C}{Ccol}{-2.1mm}{1mm}

	\factorcolor{myblue}{f1}{f1col}
	\factorcolor{myblue}{f2}{f2col}

	\draw (A) -- (f1);
	\draw (B) -- (f1);
	\draw (B) -- (f2);
	\draw (C) -- (f2);

	\node[circle, draw, right = 1.4cm of A] (A1) {$A$};
	\node[circle, draw, below = 0.5cm of A1] (B1) {$B$};
	\node[circle, draw, below = 0.5cm of B1] (C1) {$C$};
	\factor{below right}{A1}{0.25cm and 0.5cm}{270}{$\phi_1$}{f1_1}
	\factor{below right}{B1}{0.25cm and 0.5cm}{270}{$\phi_2$}{f2_1}

	\nodecolorshift{myyellow}{A1}{A1col}{-2.1mm}{1mm}
	\nodecolorshift{myyellow}{B1}{B1col}{-2.1mm}{1mm}
	\nodecolorshift{myyellow}{C1}{C1col}{-2.1mm}{1mm}

	\factorcolor{myyellow}{f1_1}{f1_1col1}
	\factorcolorshift{myyellow}{f1_1}{f1_1col2}{2.1mm}
	\factorcolorshift{myblue}{f1_1}{f1_1col3}{4.2mm}
	\factorcolor{myyellow}{f2_1}{f2_1col1}
	\factorcolorshift{myyellow}{f2_1}{f2_1col2}{2.1mm}
	\factorcolorshift{myblue}{f2_1}{f2_1col3}{4.2mm}

	\coordinate[right=0.1cm of A1, yshift=-0.1cm] (CA1);
	\coordinate[above=0.2cm of f1_1, yshift=-0.1cm] (Cf1_1);
	\coordinate[right=0.1cm of B1, yshift=0.12cm] (CB1);
	\coordinate[right=0.1cm of B1, yshift=-0.1cm] (CB1_1);
	\coordinate[below=0.2cm of f1_1, yshift=0.15cm] (Cf1_1b);
	\coordinate[above=0.2cm of f2_1, yshift=-0.1cm] (Cf2_1);
	\coordinate[right=0.1cm of C1, yshift=0.12cm] (CC1);
	\coordinate[below=0.2cm of f2_1, yshift=0.15cm] (Cf2_1b);

	\begin{pgfonlayer}{bg}
		\draw (A1) -- (f1_1);
		\draw [arc, gray] (CA1) -- (Cf1_1);
		\draw (B1) -- (f1_1);
		\draw [arc, gray] (CB1) -- (Cf1_1b);
		\draw (B1) -- (f2_1);
		\draw [arc, gray] (CB1_1) -- (Cf2_1);
		\draw (C1) -- (f2_1);
		\draw [arc, gray] (CC1) -- (Cf2_1b);
	\end{pgfonlayer}

	\node[circle, draw, right = 1.4cm of A1] (A2) {$A$};
	\node[circle, draw, below = 0.5cm of A2] (B2) {$B$};
	\node[circle, draw, below = 0.5cm of B2] (C2) {$C$};
	\factor{below right}{A2}{0.25cm and 0.5cm}{270}{$\phi_1$}{f1_2}
	\factor{below right}{B2}{0.25cm and 0.5cm}{270}{$\phi_2$}{f2_2}

	\nodecolorshift{myyellow}{A2}{A2col}{-2.1mm}{1mm}
	\nodecolorshift{myyellow}{B2}{B2col}{-2.1mm}{1mm}
	\nodecolorshift{myyellow}{C2}{C2col}{-2.1mm}{1mm}

	\factorcolor{myblue}{f1_2}{f1_2col1}
	\factorcolor{myblue}{f2_2}{f2_2col1}

	\draw (A2) -- (f1_2);
	\draw (B2) -- (f1_2);
	\draw (B2) -- (f2_2);
	\draw (C2) -- (f2_2);

	\node[circle, draw, right = 1.4cm of A2] (A3) {$A$};
	\node[circle, draw, below = 0.5cm of A3] (B3) {$B$};
	\node[circle, draw, below = 0.5cm of B3] (C3) {$C$};
	\factor{below right}{A3}{0.25cm and 0.5cm}{270}{$\phi_1$}{f1_3}
	\factor{below right}{B3}{0.25cm and 0.5cm}{270}{$\phi_2$}{f2_3}

	\nodecolorshift{myblue}{A3}{A3col1}{-4.2mm}{1mm}
	\nodecolorshift{myyellow}{A3}{A3col2}{-2.1mm}{1mm}
	\nodecolorshift{myblue}{B3}{B3col1}{-6.3mm}{1mm}
	\nodecolorshift{myblue}{B3}{B3col2}{-4.2mm}{1mm}
	\nodecolorshift{myyellow}{B3}{B3col3}{-2.1mm}{1mm}
	\nodecolorshift{myblue}{C3}{C3col1}{-4.2mm}{1mm}
	\nodecolorshift{myyellow}{C3}{C3col2}{-2.1mm}{1mm}

	\factorcolor{myblue}{f1_3}{f1_3col1}
	\factorcolor{myblue}{f2_3}{f2_3col1}

	\coordinate[right=0.1cm of A3, yshift=-0.1cm] (CA3);
	\coordinate[above=0.2cm of f1_3, yshift=-0.1cm] (Cf1_3);
	\coordinate[right=0.1cm of B3, yshift=0.12cm] (CB3);
	\coordinate[right=0.1cm of B3, yshift=-0.1cm] (CB1_3);
	\coordinate[below=0.2cm of f1_3, yshift=0.15cm] (Cf1_3b);
	\coordinate[above=0.2cm of f2_3, yshift=-0.1cm] (Cf2_3);
	\coordinate[right=0.1cm of C3, yshift=0.12cm] (CC3);
	\coordinate[below=0.2cm of f2_3, yshift=0.15cm] (Cf2_3b);

	\begin{pgfonlayer}{bg}
		\draw (A3) -- (f1_3);
		\draw [arc, gray] (Cf1_3) -- (CA3);
		\draw (B3) -- (f1_3);
		\draw [arc, gray] (Cf1_3b) -- (CB3);
		\draw (B3) -- (f2_3);
		\draw [arc, gray] (Cf2_3) -- (CB1_3);
		\draw (C3) -- (f2_3);
		\draw [arc, gray] (Cf2_3b) -- (CC3);
	\end{pgfonlayer}

	\node[circle, draw, right = 1.4cm of A3] (A4) {$A$};
	\node[circle, draw, below = 0.5cm of A4] (B4) {$B$};
	\node[circle, draw, below = 0.5cm of B4] (C4) {$C$};
	\factor{below right}{A4}{0.25cm and 0.5cm}{270}{$\phi_1$}{f1_4}
	\factor{below right}{B4}{0.25cm and 0.5cm}{270}{$\phi_2$}{f2_4}

	\nodecolorshift{myyellow}{A4}{A4col}{-2.1mm}{1mm}
	\nodecolorshift{mygreen}{B4}{B4col}{-2.1mm}{1mm}
	\nodecolorshift{myyellow}{C4}{C4col}{-2.1mm}{1mm}

	\factorcolor{myblue}{f1_4}{f1_4col1}
	\factorcolor{myblue}{f2_4}{f2_4col1}

	\draw (A4) -- (f1_4);
	\draw (B4) -- (f1_4);
	\draw (B4) -- (f2_4);
	\draw (C4) -- (f2_4);

	\pfs{right}{B4}{2.9cm}{270}{$\phi'_1$}{f12a}{f12}{f12b}

	\node[ellipse, inner sep = 1.2pt, draw, above left = 0.25cm and 0.5cm of f12] (AC) {$R(X)$};
	\node[circle, draw] (B) [below left = 0.25cm and 0.7cm of f12] {$B$};

	\begin{pgfonlayer}{bg}
		\draw (AC) -- (f12);
		\draw (B) -- (f12);
	\end{pgfonlayer}

	\node[below = 0.5cm of C2, xshift=1.5cm] (tab_f2) {
		\begin{tabular}{c|c|c}
			$C$   & $B$   & $\phi_2(C,B)$ \\ \hline
			true  & true  & $\varphi_1$ \\
			true  & false & $\varphi_2$ \\
			false & true  & $\varphi_3$ \\
			false & false & $\varphi_4$ \\
		\end{tabular}
	};

	\node[left = 0.5cm of tab_f2] (tab_f1) {
		\begin{tabular}{c|c|c}
			$A$   & $B$   & $\phi_1(A,B)$ \\ \hline
			true  & true  & $\varphi_1$ \\
			true  & false & $\varphi_2$ \\
			false & true  & $\varphi_3$ \\
			false & false & $\varphi_4$ \\
		\end{tabular}
	};

	\node[right = 0.5cm of tab_f2] (tab_f12) {
		\begin{tabular}{c|c|c}
			$R(X)$   & $B$   & $\phi'_1(R(X),B)$ \\ \hline
			true  & true  & $\varphi_1$ \\
			true  & false & $\varphi_2$ \\
			false & true  & $\varphi_3$ \\
			false & false & $\varphi_4$ \\
		\end{tabular}
	};
\end{tikzpicture}